# 0-Step Capturability, Motion Decomposition and Global Feedback Control of the 3D Variable Height-Inverted Pendulum

Gabriel García, Robert Griffin, Jerry Pratt

*Abstract* — One common method for stabilizing robots after a push is the Instantaneous Capture Point, however, this has the fundamental limitation of assuming constant height. Although there are several works for balancing bipedal robots including height variations in 2D, the amount of literature on 3D models is limited. There are optimization methods using variable Center of Pressure (CoP) and reaction force to the ground, although they do not provide the physical region where a robot can step and require a precomputation for the analysis. This work provides the necessary and sufficient conditions to maintain balance of the 3D Variable Height Inverted Pendulum (VHIP) with both, fixed and variable CoP. We also prove that the 3D VHIP with Fixed CoP is the same as its 2D version, and we generalize controllers working on the 2D VHIP to the 3D VHIP. We also show the generalization of the Divergent Component of Motion to the 3D VHIP and we provide an alternative motion decomposition for the analysis of height and CoP strategies independently. This allow us to generalize previous global feedback controllers done in the 2D VHIP to the 3D VHIP with a Variable CoP.

## I. Introduction

Balancing a robot is an important topic in Humanoid Robotics. Avoiding falling prevents us from the costs of repair or replacement of damaged components. One of the common approaches is the use of reduced-order models for the robot, using only the Center of Mass (CoM) for the model. Some classical simplifications are to keep constant the angular momentum, the height of the CoM, and the Center of Pressure (CoP), as well as assuming a flat terrain. This model is known as the Linear Inverted Pendulum Model (LIP) [1], with a push recovery analysis tool known as the Capture Point [2].

Capture Point is used mainly for balance, analysis and control, and its connection with walking is done by performing some strategies of capturability. One way where 2D works had been extended to 3D for walking is to consider the decoupled systems on each spatial component. Following this path, the concept of the Capture Point had been extended to the well-known Divergent Component of Motion (DCM) [3-6], highly used in walking, where usually the robot plans a trajectory using only this point in the space. In [7-8] Grizzle et al. perform lateral stabilization for walking by enforcing the lateral average velocity to 0 as a constraint in an optimization problem considering a decoupled lateral-sagittal system. Here, we use these decoupling ideas for the balance control of the 3D VHIP with Fixed CoP balance as multiple 2D VHIPs.

Recently, new approaches and works arise by removing some of the classical simplifications on the LIP model. Systems modeling angular momentum have been defined in [1,9], and angular momentum had been included also in the planning stage of walking using the DCM approach, considering still a linear model [10].

There are also works considering the full centroidal dynamics including height variations, but the main difficulty in these is due to the non-convexity of the equations of motion, which requires too long solve time to be implemented in hardware applications [11-13]. Relaxations on the constraints can turn the problem into a convex one [12], and it is possible also to find inner and outer bounds of the controllable states via Sum-Of-Squares [13].

A more analytical approach was given by Koolen et. al. in [14], where they found the closed form solution for the capture region using the 2D VHIP, and provide feedback control laws. In [15], considering input limits, authors show an alternative way to control the 2D VHIP using sliding mode and feedback linearization in a DCM that models height variations. However, both [13] and [14] do not cover the 3-Dimensional case and assume a constant CoP.

Ramos et al. [16] found some possible capture points on a variable-height model using an iteration algorithm based on bisection. With this, they show that the 3D case can be solved as a 2D case defined in the direction of the push under fixed CoP and angular momentum.

In [17-19] Caron et. al. remove the assumption of fixed CoP for the stabilization of the 3D VHIP. They use optimization to compute the force that must be applied to the ground and the trajectory of the CoP in the contact surface, but they do not provide the admissible physical region allowed for balance or, equivalently, *where to step*. Also, their analysis is mostly based on time and their *boundedness condition* requires first to compute trajectories, without considering the analytical approach from [14].

The first analysis on height variations strategies using components of motion was done in [20], where authors define the called Time-Varying Divergent Component of Motion which allows for height variations. That component was analyzed again in [19], using still a time-based focus.

In this work, we extend the analytical approach of height variations done in [14, 15] to the analysis of height and CoP strategies on the 3D VHIP without using time-based analysis, but using a state-space analysis. Also, based on [20] we will define the DCM for the 3D VHIP as a function of the state variables, allowing feedback control on the model. We will also show the connection with the DCM in [15]. Additionally, our work does not restrict the new DCM to a flat terrain: we only use a flat foot.

The main contributions of this paper are presented as follows:

- Provide a transformation that shows the equivalence of the 3D VHIP with a fixed CoP with its analogous 2D case for 0-step capturability

The authors are with the Florida Institute for Human and Machine Cognition, 40 S Alcaniz St, Pensacola, FL 32502, United States
E-mail: {ggarcia,rgriffin,jpratt}@ihmc.us

- Provide the necessary and sufficient conditions for 0-step capturability of the 3D VHIP with Variable CoP *a priori* (based only in the current states).
- Provide the motion decomposition of the 3D VHIP with Variable CoP into the analogous Divergent/Convergent Component of Motion and Virtual Repellent Point.
- Provide a decomposition of the 3D VHIP with Variable CoP into a 2D VHIP with Fixed CoP and a 3D LIP with Variable CoP for the analysis of height and CoP strategies respectively.
- Provide feedback control laws into the last Decomposition without performing any trajectory optimization for balance in the whole physically possible region of stabilization.

First, we will present the dynamical system we are working on, the 3D VHIP (Section II). We will then show the first contribution relate with the 3D VHIP with Fixed CoP in Section III. In Section IV we will provide the second contribution of this paper: the necessary and sufficient conditions for stability of the 3D VHIP with Variable CoP. n Section V we will study the motion decomposition and global feedback control of the 3D VHIP. Last three contributions are in this section. Finally, we will discuss the results and we will give some open problems in Section VI.

## II. 3D VHIP MODEL

In this section we will present the dynamics of the 3D VHIP Model, the most general case of the centroidal dynamics of a robot when considering constant angular momentum and a single contact

### A. Dynamical model

Considering a single planar contact to the world, the centroidal dynamics of a robot evolve as follows:

$$\begin{bmatrix} \dot{L} \\ \ddot{r} \end{bmatrix} = \begin{bmatrix} \tau_n \hat{n} + f_{gr} \times (r - r_P) \\ \frac{1}{m} f_{gr} + g \end{bmatrix} \quad (1)$$

s.t.

$$A r_P \leq b \quad (2)$$

$$f_{gr} \cdot \hat{n} \geq 0 \quad (3)$$

Where:
$f_{gr}$: Reaction force of the ground;
$L$: Angular momentum around the CoM;
$r$: Position of the CoM $[x, y, z]$;
$r_P$: Position of the CoP $[x_P, y_P, z_P]$;
$g$: Gravity vector $[0,0, -g]$;
$\hat{n}$: Normal vector to the Contact Surface;
$\tau_n$: Normal torque produced by the contact.

(2) is the geometrical representation of the Contact Surface (considering a polygonal foot). (3) is the unilateral contact restriction.

By enforcing constant angular momentum $\dot{L} = 0$ it can be shown that $\tau_n = 0$ and that $f_{gr}$ is parallel to $r - r_P$. Physically, this implies that the force is always pointing towards the CoM. Taking this into consideration we can define the following force:

$$f_{gr} = mu(r - r_P) \quad (4)$$

Where:

$m$: Mass of the robot

$u$: New control input holding:

$$u = \frac{\|f_{gr}\|}{m\|r - r_p\|} \geq 0 \quad (5)$$

(5) is the new unilateral contact condition equivalent to (3). Inserting (4) into (1) gives us the dynamics of the CoM under the assumption of constant angular momentum:

$$\ddot{r}(t) = u(r - r_P) + g, u \geq 0, A r_P \leq b \quad (6)$$

Taking each component of the vector $r$ we obtain the dynamical system of the 3D VHIP:

$$\begin{bmatrix} \ddot{x} = u(x - x_P) \\ \ddot{y} = u(y - y_P) \\ \ddot{z} = u(z - z_P) - g \end{bmatrix}; u \geq 0 \quad (7)$$

In summary, (7) represents any humanoid model considering the following assumptions:

- Balance using only one foot (single support).
- Variable CoP on a Polygonal Contact Surface (**CS**).
- Twist (yaw) torque equal to zero.
- Force goes from the CoP to the CoM.
- Enough friction on the floor for avoid slipping.

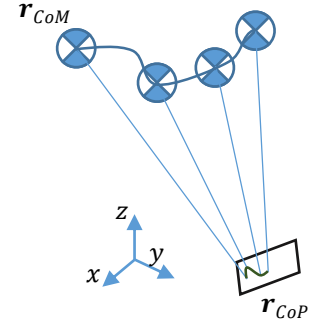

Figure 1: 3D VHIP model

The objective of balance is to drive the system given by eq. (7) to the fixed final point $(x, y, z) \rightarrow (x_{Pf}, y_{Pf}, z_f)$.

### B. Ballistic trajectory.

A special trajectory of the 3D VHIP is the *ballistic trajectory* introduced in [13], which is defined as the resultant trajectory when the robot apply a force $u = 0$. From (6) we have:

$$r_{ball}(t) = r_0 + \dot{r}_0 t + \frac{g}{2} t^2 \quad (6)$$

In other words, ballistic trajectory is the curve described by object in free fall with a given initial velocity and position. We can see in Fig. 2 the ballistic trajectory as an inverted parabola in blue.

In this paper, we will be referring by region "*below the ballistic trajectory*" as the shaded region in Fig. 2. Also, Let us define the *ballistic line* as the projection of the ballistic trajectory onto the XY plane (Plotted in green).

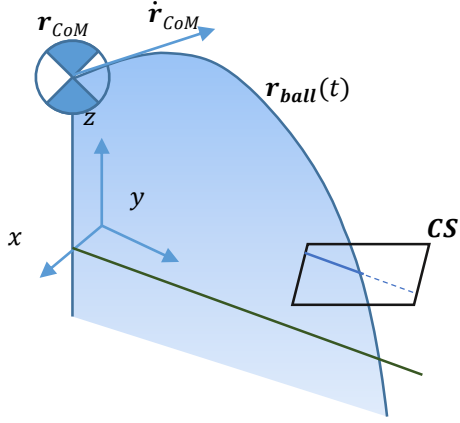

Figure 2: 3D VHIP and Ballistic trajectory

Points belonging to the line in blue in the Contact Surface $CS$ are considered "below the ballistic trajectory". Likewise, points belonging to the dashed line are considered "above the ballistic trajectory". We also say that in this case the foot has step "over the ballistic line" because its projection to the XY plane contains points of the ballistic line. The full plane containing the ballistic trajectory is the "ballistic plane".

## III. 3D VHIP FIXED COP

Models using a fixed Center of Pressure are known as "point-foot models". The system given by (7) corresponds to the 3D VHIP for any CoP. In this section, we are going to fix it to the origin.

$$(x_P, y_P, z_P) = (0,0,0) \tag{8}$$

This leads to the following dynamical system called in this subsection "3D VHIP with fixed CoP", which is very similar to the 2D VHIP showed in [14]:

$$\begin{bmatrix} \ddot{x} = ux \\ \ddot{y} = uy \\ \ddot{z} = uz - g \end{bmatrix}; u \geq 0 \tag{9}$$

In this section we will be referring as a "VHIP stable" if that VHIP can be balanced (i.e, $(x, y, z) \to (x_{Pf}, y_{Pf}, z_f)$ with convergent zero final velocity).

### A. Possible stabilization region

Let us define the *ballistic line* as the projection of the ballistic trajectory onto the XY plane.

**Lemma 1**: *If the 3D VHIP with Fixed CoP is stable, then the Center of Pressure must be placed over the ballistic line.*

*Proof:* Let us introduce the variable:

$$w = \dot{x}y - \dot{y}x \tag{10}$$

which has dynamics:

$$\dot{w} = 0 \tag{11}$$

From (11) we know that $w$ is constant (actually, it is an invariant set). We wish to have $w \to w_f = 0$, so we have $w = 0$ for all times because $w$ is constant. This implies that to achieve balance, we need:

$$\dot{x}y = \dot{y}x \ \forall\ t \geq 0 \tag{12}$$

The equality $\frac{\dot{x}}{\dot{y}} = \frac{x}{y}$ is held if and only if the XY components of the CoP (0,0) are placed on the XY direction in which the robot was pushed, i.e, the ballistic line.

□

### B. Equivalence of the stable 3D VHIP to a 2D VHIP

The dynamical system in the ballistic line can be reduced to a simple 2D VHIP and we can apply a control law like [14,15], so we can stabilize the system with already existant feedback control laws.

**Lemma 2**: *If the 3D VHIP with Fixed CoP is stable, then it is can be controlled as the 2D VHIP model*.

*Proof:* Let us decouple the system into two 2D VHIPs, one sagittal and one lateral. We are going to prove that if we met the condition of Lemma 1, i.e. the CoP is on the ballistic line, then *any* control law that stabilizes one of the two 2D VHIP will stabilize the whole system.

Let us take $(x, z)$ and $(y, z)$ as the two 2D VHIP decoupled, lateral and sagittal plane.

We will have corresponding to the lateral 2D VHIP $(x, z)$:

$$\begin{bmatrix} \ddot{x} = ux \\ \ddot{z} = uz - g \end{bmatrix}; u \geq 0 \tag{13}$$

with initial conditions:

$$x(0) = x_0, \ \dot{x}(0) = \dot{x}_0 \tag{14}$$

We have the dynamics of the sagittal 2D VHIP $(y, z)$:

$$\begin{bmatrix} \ddot{y} = uy \\ \ddot{z} = uz - g \end{bmatrix}; u \geq 0 \tag{15}$$

with initial conditions:

$$y(0) = y_0, \ \dot{y}(0) = \dot{y}_0 \tag{16}$$

If we assume the 3D VHIP is stable, then Lemma 1 holds and from (12):

$$\dot{x}y - \dot{y}x = 0$$

This equality defines the dynamics along the ballistic plane. Let us change the variable of the sagittal plane $y$ to $y_2$:

$$y_2 = \frac{x_0}{y_0} y \tag{17}$$

Then, by using (12) when $t = 0$ in the derivative of (17) we have:

$$\dot{y}_2 = \frac{x_0}{y_0} \dot{y} = \frac{\dot{x}_0}{\dot{y}_0} \dot{y} \tag{18}$$

$$\ddot{y}_2 = \frac{x_0}{y_0} uy = uy_2 \tag{19}$$

This yields the new sagittal scaled system $(y_2, z)$:

$$\begin{bmatrix} \ddot{y}_2 = uy_2 \\ \ddot{z} = uz - g \end{bmatrix}; u \geq 0 \tag{20}$$

And the initial condition holds:

$$y_2(0) = x_0, \dot{y}_2(0) = \dot{x}_0 \tag{21}$$

Note that this sagittal scaled system (20) and (21) is the same as the lateral 2D VHIP (13) and (14). Due to the existence and uniqueness theorem of differential equations, we have $y_2(t) = x(t)$, supposing that $u(t)$ is Lipschitz continuous. Although some control laws $u$ can be piecewise continuous (and not Lipschitz continous), the result of a unique solution still persist as long as $u(t)$ is not multivalued (impossible for a scalar value). We have now:

$$y_2(t) = x(t), \forall t \geq 0 \quad (22)$$

$$\lim_{t \to \infty} y_2(t) = \lim_{t \to \infty} x(t) = 0$$

Using (17) we obtain.

$$\lim_{t \to \infty} y(t) = 0 \quad (23)$$

Lemma 2 follows: If the CoP is placed on the ballistic line, and a lateral 2D VHIP is stabilized, then the sagittal scaled system and the whole system are stabilized.

□

**Corollary 1**: *The 3D VHIP with Fixed CoP only admits a 1D region as the Capture Region, and it is the intersection of the 2D region below the ballistic trajectory with the solid ground.*

Using Lemma 1 from [14], we know that the necessary and sufficient condition for balance is that the CoP must be placed below the ballistic trajectory in the 2D VHIP. Using this fact, along with Lemma 2, we assert that the CoP must be placed below the 3D ballistic trajectory, and also in the ground. Corollary 1 follows as the first contribution of this paper.

IV. 3D VHIP WITH VARIABLE CoP

In this section, we will consider the full system given by (5). We have the following dynamical system as the 3D VHIP with variable CoP:

$$\begin{bmatrix} \ddot{x} = u(x - x_P) \\ \ddot{y} = u(y - y_P) \\ \ddot{z} = u(z - z_P) - g \end{bmatrix}, u \geq 0, \boldsymbol{r_P} \in \boldsymbol{CS} \quad (24)$$

where $\boldsymbol{r_P} = (x_P, y_P, z_P)$ are the coordinates of the Center of Pressure of the foot. By allowing the CoP to vary, we can now consider $\boldsymbol{r_P}$ a control input along with $u$. $\boldsymbol{CS}$ is the contact surface where the reaction force of the ground can be applied. The restriction $\boldsymbol{r_P} \in \boldsymbol{CS}$ can be modeled as the polygonal constraint of the foot in the floor like:

$$\boldsymbol{A r_P} \leq \boldsymbol{b} \quad (25)$$

Note that the foot is a 2D surface, so one of the variables might be restricted to the other ones (e.g. $z_P = Z(x_P, y_P)$ because of the foot). In the practice we will have only 3 control variables instead of the 4. $u$ is related to the reaction force applied by the leg, called "stiffness" in some works [17-19]. From (5), controlling $u$ is done by changing the amount of force we apply to the ground.

A. *Necessary and sufficient condition for balance*

In [18], a necessary and sufficient condition for balance was given as a *boundedness condition*. The main problem with this approach is that we need *first* to find a trajectory for the normalized force $u(t)$ and the trajectory of the Center of Pressure $\boldsymbol{r_P}(t)$. Using that approach, it is impossible to know if a given state is capturable or not without solving an optimization problem to try to find the mentioned trajectories.

In [14], authors defined $T$ as the time that the ballistic trajectory takes to reach the base of the Center of Pressure, and $z_c$ as the z-axis intercept of the ballistic trajectory in the 2D VHIP with Fixed CoP model. According to them, this model can be stabilized if and only if the values of $z_c$ and $T$ holds:

$$z_c > 0 \quad (26)$$

$$T > 0 \quad (27)$$

Equations (26) and (27) holds if and only if the CoP is below the ballistic trajectory in the2D VHIP with Fixed CoP model.

In this subsection we are going to extend that result to the 3D VHIP with Variable CoP as the second contribution of this paper: The necessary and sufficient condition for balance using *only* the current states.

**Lemma 3**: *The 3D VHIP with variable CoP is 0-step capturable if and only if the Contact Surface (CS) has a non-empty intersection with the region below the ballistic trajectory.*

The proof uses invariant sets for showing that, if there is a plane $B$ separating the $\boldsymbol{CS}$ with the ballistic trajectory, then this trajectory will always be pushed away from that plane.

First we present the physical meaning of the functions involved in the proof:

$h_{m2}$: Vertical distance between the CoM and a plane $B$.
$\dot{h}_{m2}$: Positive when moving towards the plane**.**
$h_m$: Maximum vertical distance between the ballistic trajectory and the plane $B$ for a given instant.

We are going to use the compact, vector form from (6):

$$\ddot{\boldsymbol{r}} = u(\boldsymbol{r} - \boldsymbol{r_P}) + \boldsymbol{g}, u \geq 0$$

*Proof of Lemma 3*: We can define a plane $B$ below $\boldsymbol{CS}$ and over the ballistic trajectory. We have:

$$B: z = ax + by + c \quad (28)$$

We can then define a vector $\boldsymbol{n_B}$ orthogonal to $B$ and with z-component equal to 1:

$$\boldsymbol{n_B} = [-a, -b, 1]^T \quad (29)$$

By definition, any point $\boldsymbol{r_P}$ in $\boldsymbol{CS}$ is above $B$, so:

$$\boldsymbol{n_B^T r_P} > c, \forall \boldsymbol{r_P} \in \boldsymbol{CS} \quad (30)$$

The final position $\boldsymbol{r_f}$ of the CoM has the form:

$$\boldsymbol{r_f} = \boldsymbol{r_{Pf}} + h_f \hat{\boldsymbol{e}}_z \quad (31)$$

with $\boldsymbol{r_{Pf}}$ being the final position of the Center of Pressure, $h_f > 0$ the final height, and $\hat{\boldsymbol{e}}_z$ the unitary vector in the z direction.

See that $\boldsymbol{r_f}$ is also above $B$:

$$\boldsymbol{n_B^T r_f} = \boldsymbol{n_B^T r_{Pf}} + h_f \boldsymbol{n_B^T} \hat{\boldsymbol{e}}_z = \boldsymbol{n_B^T r_{Pf}} + h_f > c \quad (32)$$

We can now define the following function, which gives a condition for capturability:

$$h_{m2}(t) = \boldsymbol{n_B^T r}(t) - c \quad (33)$$

Its derivatives hold:

$$\dot{h}_{m2}(t) = \boldsymbol{n_B^T \dot{r}}(t) \quad (34)$$

$$\ddot{h}_{m2}(t) = \boldsymbol{n_B^T}(u(\boldsymbol{r}(t) - \boldsymbol{r_P}) + \boldsymbol{g})$$

$$\ddot{h}_{m2} = u(h_{m2} - w_r) - g \quad (35)$$

with:
$$w_r = -c + \mathbf{n}_B^T \mathbf{r}_P > 0 \quad (36)$$

First we will show that, if the system is capturable, we require $\dot{h}_{m2} > 0$ when $h_{m2} < 0$. We are going to show that the set $\mathcal{O}$ is invariant using the functions $V_1$ and $V_2$:

$$\mathcal{O} = \{(h_{m2}, \dot{h}_{m2}) \in \mathbb{R}^2 |\ h_{m2} < 0 \wedge \dot{h}_{m2} < 0\} \quad (37)$$

$$V_1 = h_{m2} \quad (38)$$

$$V_2 = \dot{h}_{m2} \quad (39)$$

$$\dot{V}_1 = \dot{h}_{m2} < 0, \forall (h_{m2}, \dot{h}_{m2}) \in \mathcal{O} \quad (40)$$

$$\dot{V}_2 = u(h_{m2} - w_r) - g < 0, \forall (h_{m2}, \dot{h}_{m2}) \in \mathcal{O} \quad (41)$$

We require $h_{m2}(t_f) > 0$, so if the system is inside of $\mathcal{O}$, then $h_{m2}$ will never be positive and the system would not be capturable. In consequence if $h_{m2} < 0$, then we require for capturability $\dot{h}_{m2} \geq 0$.

We can now prove Lemma 3 by considering the function

$$h_m(t) = \mathbf{n}_B^T \mathbf{r}(t) + \frac{(\mathbf{n}_B^T \dot{\mathbf{r}}(t))^2}{2g} - c \quad (42)$$

and showing that the set $\mathcal{O}_2$

$$\mathcal{O}_2 = \{h_m \in \mathbb{R} |\ h_m < 0\} \quad (43)$$

is also invariant using the function $V$:

$$V = h_m \quad (44)$$

Taking time derivative:

$$\dot{V} = \frac{\mathbf{n}_B^T \dot{\mathbf{r}}(t)}{g} u \mathbf{n}_B^T (\mathbf{r}(t) - \mathbf{r}_P)$$

$$\dot{V} = \frac{\dot{h}_{m2}(t)}{g} u(h_{m2}(t) - w_r) \quad (45)$$

We know that $h_m > h_{m2}$ (because of the square term) so in $\mathcal{O}_2$ we have $0 > h_{m2}$. This implies for capturability $\dot{h}_{m2} \geq 0$ as we derived using the set $\mathcal{O}$. Considering those inequalities along with $w_r > 0$ and $u \geq 0$ we obtain

$$\dot{V} \leq 0, \forall h_m \in \mathcal{O}_2 \quad (46)$$

but $h_m(t_f) = h_{m2}(t_f) > 0$, so the trajectories must not enter to $\mathcal{O}_2$ (because it is an invariant set). We have now that for 0-step capturability we require $h_m \notin \mathcal{O}_2$ and $(h_{m2}, \dot{h}_{m2}) \notin \mathcal{O}$ for any $\mathbf{n}_B$. We can now show why this is equivalent to Lemma 3.

If the Contact Surface $\mathbf{CS}$ does not intersect with the region below the ballistic trajectory, and taking into consideration the convexity of both of them, we can always define a plane $B$ with normal $\mathbf{n}_B$, with $\mathbf{CS}$ above $B$ and the ballistic trajectory below it. In particular, the initial position of the CoM is below $B$, so:

$$h_{m2}(0) = \mathbf{n}_B^T \mathbf{r}(0) - c < 0 \quad (47)$$

If we want capturability, then we need to avoid $(h_{m2}, \dot{h}_{m2}) \in \mathcal{O}$ when $t = 0$:

$$\dot{h}_{m2}(0) \geq 0 \Leftrightarrow \mathbf{n}_B^T \dot{\mathbf{r}}(0) \geq 0 \quad (48)$$

The ballistic trajectory as a curve parametrized by $\tau$ when $t = 0$ is defined as:

$$\mathbf{r}_b(\tau) = \left(x_0 + \dot{x}_0 \tau, y_0 + \dot{y}_0 \tau, z_0 + \dot{z}_0 \tau - \frac{g}{2}\tau^2\right)$$

$$\mathbf{r}_b(\tau) = \mathbf{r}_0 + \dot{\mathbf{r}}_0 \tau + \frac{g}{2}\tau^2, \forall \tau \geq 0 \quad (49)$$

The vertical distance from the ballistic trajectory when $t = 0$ to the plane B is:

$$h(\tau) = \mathbf{n}_B^T \mathbf{r}_b(\tau) - c$$

$$h(\tau) = \mathbf{n}_B^T \mathbf{r}_0 + \mathbf{n}_B^T \dot{\mathbf{r}}_0 \tau - \frac{g}{2}\tau^2 - c \quad (50)$$

We have supposed that the whole ballistic trajectory is below the plane B, so we have:

$$h(\tau) < 0, \forall \tau \geq 0 \quad (51)$$

In particular, the closest point of the ballistic trajectory to the plane (in vertical distance) is still below $B$. This closest point corresponds to the maximum of $h(\tau)$, defined when:

$$\tau_m = \frac{\mathbf{n}_B^T \dot{\mathbf{r}}(t)}{g} = \frac{\dot{h}_{m2}(0)}{g} > 0 \quad (52)$$

$$h(\tau_m) < 0 \quad (53)$$

$$h(\tau_m) = \mathbf{n}_B^T \mathbf{r}_0 + \frac{(\mathbf{n}_B^T \dot{\mathbf{r}}_0)^2}{2g} - c$$

$$h(\tau_m) = h_m(0) < 0 \quad (54)$$

We have now that $h_m(0) < 0 \Leftrightarrow h_m \in \mathcal{O}_2$ which contradicts the requirement for capturability $h_m \notin \mathcal{O}_2$. In consequence, there must not exist any plane $B$ dividing the region below the ballistic trajectory and the Contact Surface $\mathbf{CS}$. This implies finally that those regions must always intersect as a necessary condition of capturability.

Now let us see why this is sufficient for capturability. You can always find a point $\mathbf{r}_{fP} \in \mathbf{CS}$ such that the ballistic trajectory is over it. The z-intercept of the ballistic trajectory in a frame of coordinates centered on $\mathbf{r}_{fP}$ will be positive, and also $\mathbf{r}_{fP}$ is forward in the direction of the initial velocity. These statements are condensed into the following two inequalities in the new frame, recapping (26) and (27):

$$z_c > 0$$

$$T > 0$$

From [14], this is enough for balance, so the system is controllable. We can apply a control law based on Orbital Energy and the system will be stable because (26) and (27) are held. This shows sufficiency of Lemma 3.

□

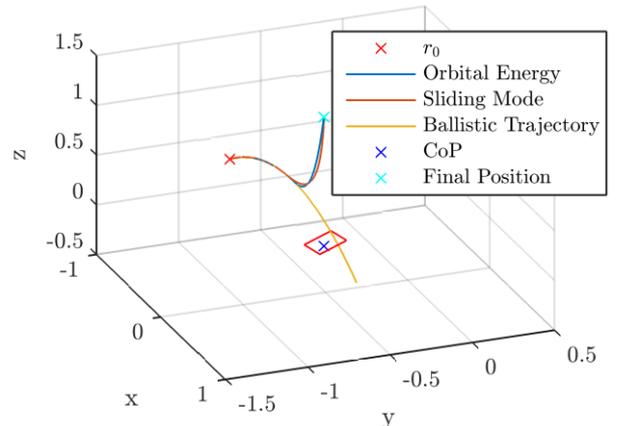

Figure 3: Trajectories obtained using feedback control laws showed in [14] and [15].

Fig. 3 shows an example of the sufficiency of Lemma 3. We chose a $r_{fP} \in CS$ below the Ballistic Trajectory, and we apply the controllers showed in [14] and [15] by just fixing the CoP at $r_{fP}$. Although this is not using variations on the CoP, it is enough to illustrate sufficiency of the Lemma 3.

Obviously, fixing the CoP is only one way to control the robot, and it is not considering the possible advantages of use a variable CoP. In Section V we will show a way to control in *feedback* form the 3D VHIP, without using any time function or open loop control.

We will now highlight some corollaries consequences of Lemma 3.

**Corollary 2**: *For 0-step capturability of the 3D VHIP with Variable CoP, the foot must have a non-empty intersection with the ballistic line in the plane.*

If the foot does not have an intersection with the ballistic line in the plane $(x, y)$ then Lemma 3 applies. The foot is a connected set, for this to occur, it must be fully on one side of the line, as shown in Fig. 4.

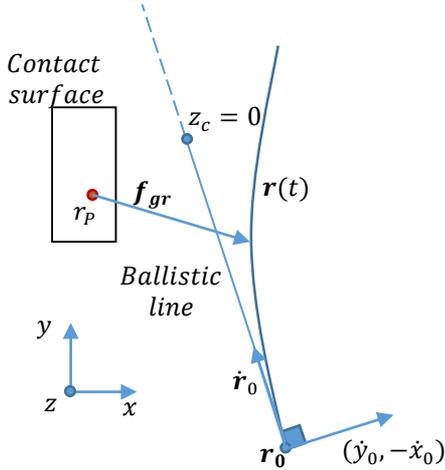

Figure 4: Contact surface of the robot fully on one side of the ballistic line.

We can also intuitively realize that the CoM is pushed away from the ballistic line, because the force applied will always be in the opposite direction of the side where the foot is, so the robot will enter to the other side and we will be unable to drive it to the contact surface.

**Corollary 3**: *When the foot is placed over the ballistic line projected to the XY plane, but above the ballistic trajectory, the system is uncontrollable.*

Fig. 5 shows an example of this case. In terms of (27) and (28), the condition $z_c > 0$ (CoP above the ballistic trajectory) can never be achieved in any point in $CS$. Because there exists a plane separating the foot and the ballistic trajectory Lemma 3 cannot be hold as the ballistic trajectory will be always pushed away from that plane. This means then that the CoM will never go over the Contact Surface.

**Corollary 4**: *In the 3D VHIP with variable CoP, the contact surface must have at least one point placed forward in the direction of the push.*

This is the analogous condition of the requirement for the 2D VHIP, $T = -\frac{x}{\dot{x}} > 0$ showed in [14]. Although in 2D the definition of "direction of the push" is simple (forward or backward, in 3D coordinates XY form a plane and the direction of the push is a line.

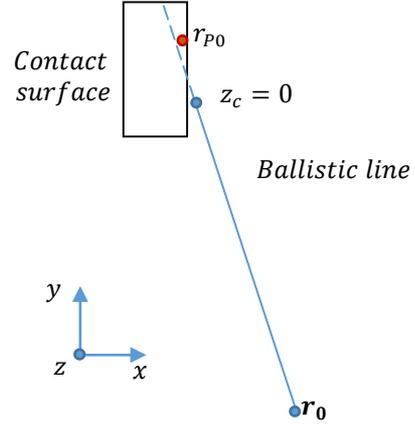

Figure 5: Trajectory of the CoM $r(t)$. Any control law on $r_P(t)$ and scaled reaction force $u$ is unable to balance the CoM

### B. Small Extension to 1-step capturability

Although this paper is focused primarily in 0-step recovery, there is one implication that can be useful for the last step in the 1-step recovery. We have considered the contact surface $CS$ as the region where you can place the CoP as desired (in single support, it is only one foot). In double support you can place the CoP in any place of the polygonal support or the convex hull of the projection to the XY plane of the feet, so this will be our new Contact Surface. Note that for the proof of *Lemma 3* we did not used the shape of the terrain: it can be flat, discontinuous or roughly. Also, we have not used a planar $CS$, *Lemma 3* is applicable to *any* convex region $CS$, obtained for example, as the geometrical outer approximation of the region where you can place a point of the more general Centroidal Momentum of Pressure in a Multi-contact 3D VHIP with non-coplanar points.

**Corollary 5**: *Suppose the last step of a 1-step recovery is coplanar with the first contact. The system is controllable if and only if the convex hull of the feet contains at least one point below the ballistic trajectory.*

In this case, we no longer require the foot stepping over the ballistic line, but we require the new Contact Surface i.e. the convex hull of the feet (also called support polygon), containing the ballistic line as Fig. 6 shows.

Three non-valid steps are shown in Fig. 6, the one in the top left represents a step failing Corollary 2, the Contact Surface does not cross the ballistic line. The step in the top right is not holding Corollary 3 because the ballistic line crosses the support region but does not contain a point above the ballistic trajectory. The invalid step of the bottom fails Corollary 4, although its contact surface crosses the projection of the ballistic line, it is placed "forward" the transversal line of the initial conditions. The solid red line foot placement meets conditions of Lemma 3, so the system can be balance.

Note that the previous placements were based on the classical Instantaneous Capture Point (ICP). This time the ICP is not playing an important role here, *but* it will be a good indicator *where to step*. Our reference for the conditions for balance this time is the point where the ballistic trajectory crosses the ground. These statements will be detailed in Section V.

Once a stable step has been planned (like the red continuous line one), a control law based on the 2D VHIP can be performed. In this case we can proceed to fix the CoP in the path of the ballistic line and apply a feedback control law like [14] or we can also perform an optimization-based controller like [18] (with the drawback of getting a time-based controller). $r_{Pff}$ a fixed CoP possibility for a 2D VHIP.

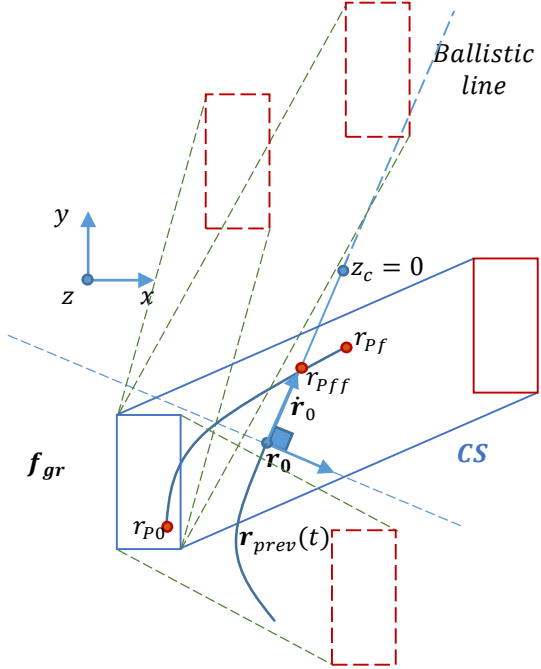

Figure 6: Last step of a 1-step recovery in red. Red steps in segment lines represents non-valid steps along with their contact surface in green segmented lines.

## V. Motion Decomposition and Control

In this section, we will limit our discussion to the 3D VHIP with Variable CoP. We will decompose the 3D VHIP into Components of Motion. The basis of this section is the Time-varying Divergent Component of Motion presented on [20], but we instead will define the variables as *virtual states* instead of time-based values.

We will follow the *based-on-states* approach used in the 2D VHIP from [14] and [15] where the variables $T$ and $z_c$ are used for getting the necessary and sufficient conditions for balance.

### A. Instantaneous curves and connections with 3D LIP and 3D DCM

In this subsection we will present two important curves which are related to the stabilization of the 3D VHIP; namely the IBT and the ICC.

We can define the Instantaneous Ballistic Trajectory (IBT) similar to Eq. (6) from [14] as:

$$r_{IBT}(\tau) \coloneqq r + \dot{r}\tau + \frac{g}{2}\tau^2, \tau > 0 \quad (55)$$

We also define the Instantaneous Capture Curve (ICC) as:

$$r_{ICC}(\tau) \coloneqq r + \dot{r}\tau + g\tau^2, \tau > 0 \quad (56)$$

Additionally, we define the Instantaneous Divergent Curve (IDC) as:

$$r_{DCM}(\tau) \coloneqq r + \dot{r}\tau \quad (57)$$

Note that the classical ICP is the crossing between the curve $r_{ICC}(\tau)$ and the ground.

The most used and common case is when $\dot{z} = 0$: Here, we can define a constant:

$$\tau_0 = \sqrt{\frac{z - z_p}{g}} = \sqrt{\frac{\Delta z}{g}}$$

And, by defining $\omega_0 = \frac{1}{\tau_0}$, we can see the relation between the ICC and *any* ICP.

$$r_{ICC}(\tau_0) \coloneqq \begin{bmatrix} x + \frac{1}{\omega_0}\dot{x} \\ y + \frac{1}{\omega_0}\dot{y} \\ z_p \end{bmatrix}$$

We can see that the ICP for the constant $\omega_0 = \sqrt{\frac{g}{\Delta z}}$ is placed at $r_{ICC}(\tau_0)$ assuming ground at height $z_p$. So, basically any ICP is a point on the ICC depending on the ground height.

Works focused on the ICP are based on the control of $r_{ICC}(\tau_0)$ as a point on the ground and its dynamics are linear when a constant $u = \omega_0^2$ is applied. Works using the 3D-DCM control for some value $\tau$ the point $r_{DCM}$ belonging to $r_{IDC}(\tau)$:

$$r_{DCM}(\tau) = r + \dot{r}\tau$$

The value of $\tau$ is a design variable chosen by the user. In [5], authors use

$$\tau = b = \sqrt{\frac{\Delta z}{g}}$$

where $\Delta z = z - z_p$ is an arbitrary constant and $z_p$ is the height of a virtual plane. Note that, although the value $\tau$ is no longer forced to be $\frac{1}{\omega_0}$ (i.e. the time constant for the ICP respect to the real ground as the approaches using the 3D LIP), $\tau$ is forced to be constant, and the value of $u$ is also forced to be constant according to Eq. (6) from [5].

### B. Generalization of the DCM to the 3D-VHIP

We desire to use a DCM for a completely variable $u$, which produces the 3D VHIP model, so we need to redefine the election of $\tau$. The DCM is special because the CoM dynamics are attracted by it, it is repelled by a vertical projection of the CoP (point called Virtual Repellent Point (VRP)) and its dynamics are independent of the CoM. A $\tau$ constant produces a 3D DCM because, so for a complete Variable-Height model, $\tau$ must be variant. This is the same analysis done in [20] with the function $\omega(t)$, but we are using $\tau$.

The new $\xi$ is any point on $r_{DCM}(\tau)$ when $\tau$ is allowed to vary.

$$\xi \coloneqq r_{DCM}(\tau) = r + \dot{r}\tau \quad (58)$$

Its dynamics using (6) are:

$$\dot{\xi} = u\tau\left(\xi + \dot{r}\left(\frac{\dot{\tau}+1}{u\tau} - \tau\right) - \left(r_P - \frac{g}{u}\right)\right)$$

If we want the dynamics of $\xi$ to be independent of the CoM, then we need to force:

$$\frac{1}{u\tau}(\dot{\tau}+1) - \tau = 0 \Leftrightarrow \dot{\tau} = -1 + \tau^2 u \quad (59)$$

Note the similarity of (59) with the called Time-Varying DCM $\omega(t)$ from [20]. The variables actually are equivalent (i.e. $\tau = \frac{1}{\omega}$), but we prefer to use $\tau$ because of its physical meaning (virtual time parametrizing the IBT, the ICC, and the IDC). Later we will see its connection to the DCM of the 2D VHIP with Fixed CoP shown in [15].

We can now define also a point dependent on the control input $u$ called the generalized Virtual Repellent Point (gVRP), which is a vertical projection of the Center of Pressure:

$$r_{gVRP} = r_P - \frac{g}{u} \quad (60)$$

using this point, we have the new dynamics of $\xi$ assuming (59):

$$\dot{\xi} = u\tau(\xi - r_{gVRP}) \quad (61)$$

Equation (61) will help us to define the generalized DCM for the 3D VHIP in Subsection D.

*C. 2D-VHIP with Fixed CoP and DCM*

In [15] the variables $T$ and $z_c$ were introduced as part of the Divergent Component of Motion. Their definitions are in (26) and (27). The DCM of the 2D-VHIP with Fixed CoP holds:

$$\dot{T} = -1 + T^2 u \quad (62)$$

$$\dot{z}_c = uT\left(z_c - \frac{g}{2}T^2\right) \quad (63)$$

This transformation shines when controlling the 2D VHIP with Fixed CoP (at the origin), mainly because the dynamics of $T$ are autonomous over itself. The introduction of a variable CoP breaks the autonomy of (62) when $T$ is defined as (27). Its new dynamics are dependent on the position of the CoM and the CoP.

*D. Augmented 3D VHIP with Variable CoP and gDCM/gCCM Decomposition*

Let us note the similarity between (59) and (62). If we were able to produce a function of the states $T = T(r, \dot{r})$ holding (62), then we could define the DCM as $\xi := r + \dot{r}T(r, \dot{r})$. But there is no function $T(r, \dot{r})$ holding (62), mainly because the introduction of the variable CoP..

Motivated by (59) and [20], we propose to create a virtual state $T_g$ ($T$ generalized) that holds the same dynamics of (62) and:

$$\dot{T}_g = -1 + T_g^2 u, \quad T_g(0) = T_{g0} \quad (64)$$

We will explain the initialization of $T_{g0}$ in subsection $F$. Taking (64) along with (6) we have the Augmented 3D VHIP with Variable CoP, with 7 scalar state variables $(T_g, r, \dot{r})$:

$$\begin{bmatrix} \dot{T}_g = -1 + T_g^2 u \\ \ddot{r}(t) = u(r - r_P) + g \end{bmatrix}, u \geq 0 \quad (65)$$

Because $T_g$ holds the same dynamics of (59), we propose to use the following change of variable for obtaining the gDCM (Generalized Divergent Component of Motion):

$$\xi_g = r + \dot{r}T_g \quad (66)$$

The new dynamics of the Augmented 3D VHIP with Variable CoP are then:

$$\begin{bmatrix} \dot{T}_g = -1 + T_g^2 u \\ \dot{\xi}_g = uT_g(\xi_g - r_{gVRP}) \\ \dot{r} = \frac{1}{T_g}(\xi_g - r) \end{bmatrix}, u \geq 0 \quad (67)$$

With $r_{gVRP}$ defined in (60). We can confirm that these dynamics are the generalization of the 3D-DCM because as long as $T_g$ is positive, the gDCM is repelled by the gVRP and the CoM is attracted by the gDCM.

*E. Augmented 3D VHIP with Variable CoP and Height/CoP Strategy Decomposition*

In the following subsections, the analysis will be done using a horizontal foot $z_p = 0$. In the last subsection we will show how to transform *any* flat foot orientation to a horizontal one, so the results are still preserved. Simulations have been done using a non-horizontal foot contact.

Given the importance of the Ballistic Trajectory as the perfect modeler of the necessary and sufficient conditions for balance, instead of controlling the gDCM we propose to control the Ballistic Trajectory.

In particular, we are going to control an important point of the $r_{IBT}$ from (55). We define the *Critical Ballistic Point CBP* as the point $\phi$ in the ballistic trajectory when $\tau = T_g$ defined by (64):

$$\phi := r_{IBT}(T_g) = r + \dot{r}T_g + \frac{g}{2}T_g^2 \quad (68)$$

Note that this leads to the relation:

$$\phi = \xi_g + \frac{g}{2}T_g^2 \quad (69)$$

and in particular:

$$\phi_{xy} = \xi_{gxy} \quad (70)$$

$$\phi_z = \xi_{gz} - \frac{g}{2}T_g^2 \quad (71)$$

We can see that the *x-y* components of $\phi$ are the same as the *x-y* components of $\xi_g$, but the *z* component of $\phi$ is shifted $\frac{g}{2}T_g^2$ units downwards from $\xi_g$. We will call the z-component of $\phi$, $z_{cg}$ (z-critical generalized). This is because it holds the same dynamics as the variable *z-critical* $z_c$ from [15], as we will see later in (75).

The dynamics using $\phi$ hold:

$$\begin{bmatrix} \dot{T}_g = -1 + T_g^2 u \\ \dot{\phi} = uT_g\left(\phi - r_p + \frac{g}{2}T_g^2\right) \\ \dot{r} = \frac{1}{T_g}(\phi - \frac{g}{2}T_g^2 - r) \end{bmatrix}, u \geq 0 \quad (72)$$

The dynamics presented on (72) can then be split into 3 subsystems. The first with $T_g$ and $z_{cg}$ ($\phi_z$), the second with $\xi_{gxy} = \phi_{xy}$, and the third with the CoM position $r$:

$$\begin{bmatrix} \dot{T}_g = -1 + T_g^2 u \\ \dot{z}_{cg} = uT_g\left(z_{cg} - \frac{g}{2}T_g^2\right) \end{bmatrix} \quad (73)$$

$$[\dot{\xi}_{gxy} = uT_g(\xi_{gxy} - r_{pxy})] \quad (74)$$

$$\begin{bmatrix} \dot{r}_{xy} = \frac{1}{T_g}(\xi_{gxy} - r_{xy}) \\ \dot{z} = \frac{1}{T_g}(z_{cg} + \frac{g}{2}T_g^2 - z) \end{bmatrix} \quad (75)$$

For the dynamics of $z_{cg}$ we have used the initial supposition $z_p = 0$. From [15], subsystem (73) corresponds *exactly* to the DCM of the 2D VHIP Fixed CoP. From [5], subsystem (74) corresponds to the *x-y* components of the DCM of the 3D LIP Variable CoP when $u$ is constant. This means approaches used in the classical 3D-DCM works based on CoM strategies can be used. Finally, subsystem (75) corresponds to the stable dynamics of the CoM. We will call (75), the Generalized Convergent Component of Motion (gCCM).

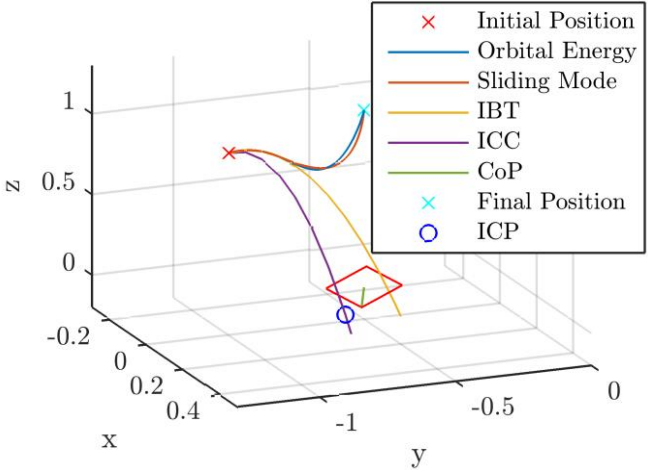

Figure 7: Balance of the 3D VHIP using the decoupled height and CoP Strategies in subsystems (73) and (74) respectively.

Although intuitively is more useful to control $\xi_{gz}$ instead of $z_{cg}$ (because z tracks directly $\xi_{gz}$ instead of $z_{cg}$ from (75)), the dynamical system (73) is rewritten as:

$$\begin{bmatrix} \dot{T}_g = -1 + T_g^2 u \\ \dot{\xi}_{gz} = uT_g\left(\xi_{gz} - \frac{g}{u}\right) \end{bmatrix} \quad (76)$$

We prefer the use of subsystem (73) instead of (76) because the control design is clearer over the variables $T_g$ and $z_c$ than in $T_g$ and $\xi_{gz}$, and because $z_{cg}$ model the necessary and sufficient condition for balance of subsystem (73) *better* than $\xi_{gz}$ ($z_{cg} > 0$ against its equivalent $\xi_{gz} > \frac{g}{2}T_g^2$).

The structure of the subsystems (73) and (74) gives us an idea of the form that the controller will take. $u$ will control $T_g$ and $z_c$ similar to the works [14,15] (using height strategies), meanwhile $r_{pxy}$ will control $\xi_{gxy}$ (using CoP strategies). Note that $r_{pxy}$ does not affect directly (73), and $u$ only affects the rate of divergence (or convergence when $r_{pxy}$ is controlling $\xi_{gxy}$ using feedback) of $\xi_{gxy}$ in (74). We can affirm now that the system is decoupled in a very convenient way for performing feedback control.

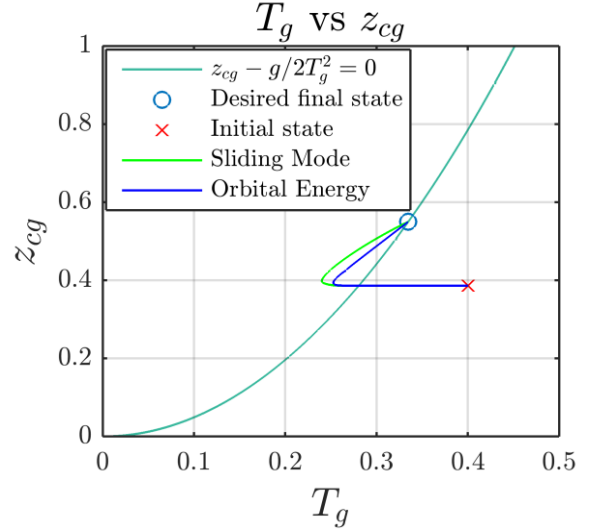

Figure 8: Subsystem $T_g$ and $z_{cg}$ for height strategies. Sliding Mode Controller from [15] and Orbital Energy Controller from [14].

### F. Initialization of variables and control

The selection of the initial state of $T_g$, $T_{g0}$, is an important stage in the control design. This will define the initial positions of the gDCM and the CBP.

In [15] authors design feedback control laws for system. Also in [14] a feedback control law was designed over the variables $(a, b)$ which are actually a transformation of $(T_g, z_{cg})$. In both cases $T_{g0}$ should be initialized as $-\frac{x_0}{\dot{x}_0}$. But this time there is no $T_{g0} = T_g(r_0, \dot{r}_0)$, as we have seen in subsection D. Because of this, we can initialize $T_{g0}$ almost *anywhere*.

Remember that $\xi_g$ attracts $r$ and it is repelled from $r_{gVRP}$ as long as $T_g$ is positive. If eventually $T_g$ becomes negative, then the dynamics are inverted: $\xi_g$ will repel $r$ and it will be attracted to $r_{gVRP}$. Furthermore it is easy to see that $T_g < 0$ is an invariant set, because in the boundary $T_g = 0$, we always have $\dot{T}_g = -1$. So, if $T_g$ becomes negative, then it will never be positive again. The same situation happens with $z_{cg}$ considering $T_g > 0$: if $z_{cg} < 0$, then $\dot{z}_{cg} < 0$ according to (73). This is equivalent to Lemma 1 from [14] in the equivalent 2D VHIP with Fixed CoP.

If $T_g$ becomes negative the system can still be controlled, but if we use the negative value of $T_g$ then we should take into account the inversion of the dynamics. The consequence is that we need to use all subsystems for control (73), (74) and (75), which goes against the use of the DCM as the unstable part of the Dynamics and the CoM as the stable part. For that reason, we will disregard the case $T_g < 0$.

We are going to initialize the variable $T_g$ such that the subsystems (73) and (74) can be controlled. When $T_g > 0$, subsystem (74) can be controlled if and only if $\xi_{gxy} = \phi_{xy}$ is inside the Contact Surface $CS$. Otherwise it will be pushed away from the foot, which corresponds with the works done on the classical 3D-DCM Linear model.

So, we need an initial value $T_{g0} > 0$ that produces an initial CBP ($\phi_0$) via $\phi_0 = r_{IBT}(T_{g0})$ where:

$$\phi_{xy0} \in CS_{xy} \qquad (77)$$

$$\phi_{z0} > 0 \qquad (78)$$

Equivalently, in terms of the gDCM using (69) we have the requirements:

$$\xi_{gxy0} \in CS_{xy}$$

$$\xi_{gz0} > \frac{g}{2}T_{g0}^2$$

Let us note how (77) and (78) remind us the necessary and sufficient conditions for balance given by Lemma 3. If the $CS$ is below the ballistic trajectory, we know we can always find a point on $r_{IBT}(\tau)$ holding (77) and (78).

Both equations impose boundaries on the one-dimensional variable $T_{g0}$ (in the beginning, $r_0$ and $\dot{r}_0$ are fixed). Because $CS$ is a convex region and using (77), there are two points $\tau$ where $\phi_{xy}(\tau)$ intercepts the boundary $CS$ in the x-y plane, which we will call $\tau_1$ and $\tau_2$ (with $\tau_2 > \tau_1$). (78) imposes a maximum value on $T_{g0}$ which is the time that the ballistic trajectory takes to cross the ground (or the extension of the plane containing the foot). We will call it $\tau_{crit}$. If $T_{g0} > \tau_{crit}$ the ballistic trajectory will be below the ground and (78) will not be hold. Putting all together along with $T_{g0} > 0$ we have bounds on $T_{g0}$:

$$\max(\tau_1, 0) < T_{g0} < \min(\tau_2, \tau_{crit}) \qquad (79)$$

An interesting value for $T_{g0}$ is $\tau_{ICP}$, which is defined as the time that takes the ICC from (56) to cross the ground. As noted previously, this crossing point is commonly referred as ICP.

The initialization $T_{g0} = \tau_{ICP}$ has a special property: using that value, we no longer require any height variations to stabilize the system (we can use a 3D LIP), or we require only linear height variations for balance (using the 3D DCM).

This suggests that a good value for initialization of $T_{g0}$ is $\tau_{ICP}$ when we desire to not use a lot of height variations. This is true when $\dot{z} = 0$: in that case, we can keep a constant height and achieve balance. When $\dot{z} \neq 0$, the system can be balance following a linear CoM trajectory. When a final desired height is specified, then height variations will be needed, but the initialization $T_{g0} = \tau_{ICP}$ is still highly reasonable for this first approach. If $\tau_{ICP}$ is outside the bounds (79), then we can choose a close value to it. Let us define some conservative bounds as:

$$\tau_{min} = k \min(\tau_2, \tau_{crit}) + (1-k)\max(\tau_1, 0) \qquad (80)$$

$$\tau_{max} = (1-k)\min(\tau_2, \tau_{crit}) + k\max(\tau_1, 0) \qquad (81)$$

for any $k$ holding $0.5 > k > 0$. $k = 0.05$ is a reasonable value. We can finally provide a good initialization of $T_{g0}$ as a saturation of $\tau_{ICP}$ between $\tau_{min}$ and $\tau_{max}$:

$$T_{g0} = \max(\tau_{min}, \min(\tau_{max}, \tau_{ICP})) \qquad (82)$$

Note again that the value $\tau_{ICP}$ can be outside the bounds of $CS$ and we can still hold (77) and (78): We can stabilize the robot even if the classical ICP is outside the support region.

Equations (77) and (78) not only define the initial condition of the variable $T_{g0}$, but the existence of a $T_{g0}$ also define the necessary and sufficient conditions for stability. Furthermore, those conditions should be hold for all times. In consequence, as mentioned in last paragraph of subsection E, the objective of $u$ is to keep both, $T_g$ and $z_{cg}$, positive in (73). Meanwhile $r_p$ should control the x-y components of $\phi$ and keep them inside $CS$ in (74).

$$r_{pxy} = \xi_{gxy} + \frac{1}{uT_g}\left(-k_{p1}(\xi_{gxy} - \xi_{gxyd})\right) \qquad (83)$$

$$r_{pxy} = \xi_{gxy} - k_{p2}(\xi_{gxy} - \xi_{gxyd}) \qquad (84)$$

Control law (83) produces an exact exponential decay on the variable $\xi_{gxy}$ towards the desired final point $\xi_{gxyd}$. Eq. (84), produces a decay in the same direction, but we cannot guarantee exponential decay. Nevertheless, it is enough to see that the term $uT_g$ is always positive, so, $\xi_{gxy}$ always approaches $\xi_{gxyd}$. Note that we also should bound $r_{pxy}$. It should hold $Ar_P \leq b$, so, in case of saturation, we should take the maximum value of $k_{p1}$ and $k_{p2}$.

The control input $u$ can be defined as the Clipped Orbital Energy Controller from [14]:

$$a = -\frac{1}{T_g}$$

$$b = \frac{1}{T_g}\left(z_{cg} + \frac{g}{2}T_g^2\right)$$

$$U(T_g, r, \dot{r}) = -7a^2 + \frac{3z_f a^3 - ga}{b} - \frac{10a^3 b}{g}$$

$$u = \max(U(T_g, r, \dot{r}), 0) \qquad (85)$$

Stability is ensured by the use of Cylindrical Algebraic Decomposition (CAD) in the whole region of stabilization $T_g > 0$. This is a result of the dynamic decomposition (73) and (74), which shows a perfectly decoupled system for almost-independent height and CoP strategies. (73) shows the exact dynamical system controlled in [14] and [15], and controller (85) is intrinsically controlling only the phase plane $T - z_c$ (equivalently, the plane $a - b$), meanwhile variables $x$ and $z$ will remain bounded as proved in [15] using Input-to-State Stability. Another possible controller considering an upper bound on control law $u$ can be defined using Sliding Mode control for (73) as:

$$u = \max(\min(U(T, z_c), 1), 0) \qquad (86)$$

This is Eq. (30) from [15]. There, authors use a reduced system and a transformation in order to use $g = 1$ and $u_{max} = 1$. The equivalence in the variables used the present work is done using an inverse transformation:

$$T_g = \frac{1}{\sqrt{u_{max}}}T \qquad (87)$$

$$z_{cg} = \frac{g}{u_{max}} z_c \quad (88)$$

So the control law to use in the frame of this 3D VHIP will be:

$$u = u_{max} \max\left(\min\left(U\left(\sqrt{u_{max}}T_g, \frac{u_{max}}{g}z_{cg}\right), 1\right), 0\right) \quad (89)$$

Where $U(\cdot,\cdot)$ is a function of the states given by (29) from [15]. Note that the final height also changes, in the new frame. We should replace all $z_f$ for $z_f \frac{u_{max}}{g}$. Note that the denormalization (87) and (88) changes the necessary and sufficient condition under upper saturation on $u$ obtained in [15] from $T > 1$ and $z_c > \frac{1}{2}$ to:

$$T_g > \frac{1}{\sqrt{u_{max}}} \quad (90)$$

$$z_{cg} > \frac{g}{2u_{max}} \quad (91)$$

Note that when $u_{max} \to \infty$, (90) and (91) are equivalent to (26) and (27). If the upper bound $u \le u_{max}$ is enforced, then the necessary and sufficient conditions of Lemma 3 modeled by (26) and (27) stops holding. The new necessary and sufficient conditions for balance of the 3D VHIP switches to the existences of a point $T_g$ where (90) and (91) hold. This is translated in spatial terms into the existence of a point of the foot below a *shifted ballistic trajectory* given by (90) and (91), plotted in Fig. 3 of [15].

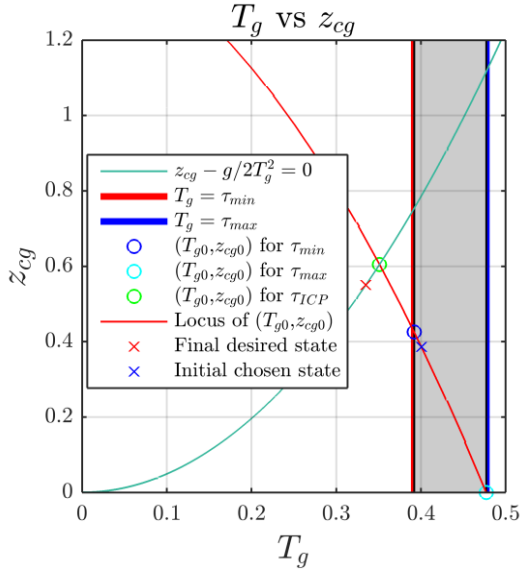

Figure 9: Initialization of $T_{g0}$ in the phase plane $T_g - z_{cg}$. Allowed values imposed by Contact Surface are shaded. We pick a conservative value on the initial conditions.

### G. Disturbance rejection and re-initialization.

In this subsection we will shortly talk about how to *reinitialize* the virtual variable $T$ in case of a high disturbance that produces the a reposition of the IBT outside the support polygon.

We can realize that the *stability* is modeled by (77) and (78). For now, we consider a *disturbance* as an instantaneous change of the state variables. In that sense, an *impulsive push* changes instantaneously the velocity of the CoM. The critical variables for stability are $\phi$ and $T_g$, or equivalently, $\xi_{gxy}$, $z_{cg}$ and $T_g$. In theory the variable $T_g$ can never be disturbed, because it is a virtual state running on the controller, but integration errors can occur, which we will ignore.

So the variables $\xi_{gxy}^-$, $z_{cg}^-$ will instantaneously change after an impulsive push. If the new $\xi_{gxy}^+$ is outside $CS$ or the new $z_{cg}^+$ is less than 0, according to (77) or (78) respectively we will not be able to stabilize the system *with* the current value of $T_g^-$: The task is then to find a new $T_g^+$ such that (77) and (78) are hold again. Note that we are able to do that because we have total authority on $T_g$, as it is only a virtual state variable with no more physical meaning than just an arbitrary time. We should recalculate $\tau_1$, $\tau_2$, $\tau_{crit}$, $\tau_{ICP}$ and reapply (82) for getting the new $T_g^+$, and apply the same feedback controllers (84) and either, (85) or (86), to the system.

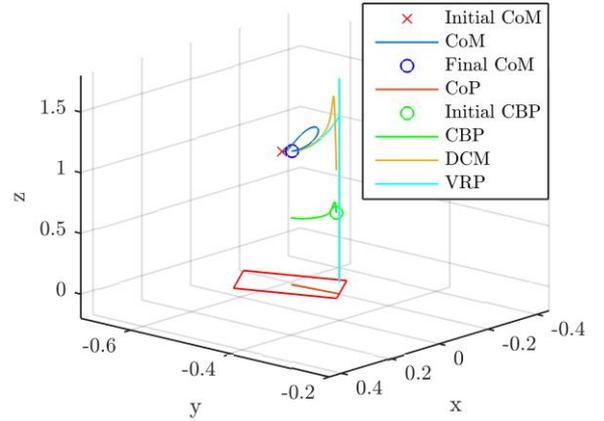

Figure 10: Balance of the 3D VHIP, DCM, VRP and CBP.

Fig. 10 shows an instantaneous change in the velocity of the center of mass of the robot. From rest, the center of mass has been changed an amount of $\Delta \dot{r} = [-0.01, 0.4, -0.2]m/s$ and the classical ICP is placed outside the foot, in a reachable time of $\tau_{ICP} = 0.319$, but the time when the CoM leaves the foot is $\tau_{max} = 0.2818$ (this is the maximum value that the DCM can take for stability). Classical strategies like 3D LIP/DCM are not able to recover without taking a step, so a height or angular momentum variation is needed. We chose a conservative reinitialization value given by (82) of $T_{g0}^+ = 0.2677$.

Fig. 11 shows the phase plane $T_g$ and $z_{cg}$, because the system was in stability and in rest, $T_g^-$ was in the stable curve $z_{cg} = \frac{g}{2}T_g^2$ showed in green. After the impulsive push is applied resulting in the ICP outside the foot, we must change the value of $T_g$ to the allowed shaded region. The locus of $T_g$ and $z_{cg}$ is a parabola because of the definition of $z_{cg}$ as the third component of (69):

$$z_{cg}^+ = z + \dot{z}T_g^+ - \frac{g}{2}T_g^{+2}$$

Note that the only decision variable is $T_g^+$, as $z$ and $\dot{z}$ are instantaneously fixed. Fig. 12 shows the Orbital Energy and the Sliding Mode Controllers after the new selection of $z_{cg}$.

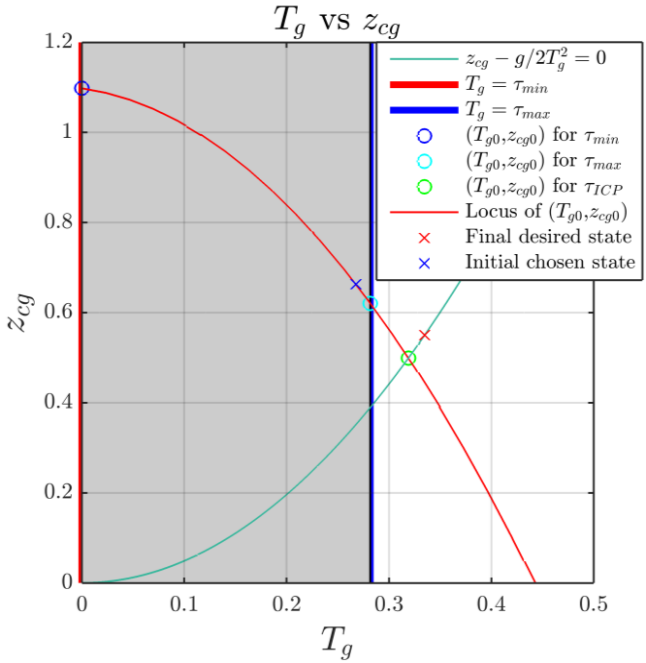

Figure 11: Re-initialization of $T_{g0}$ in the phase plane $T_g - z_{cg}$. This is necessary when a disturbance places the DCM outside of the Support Polygon.

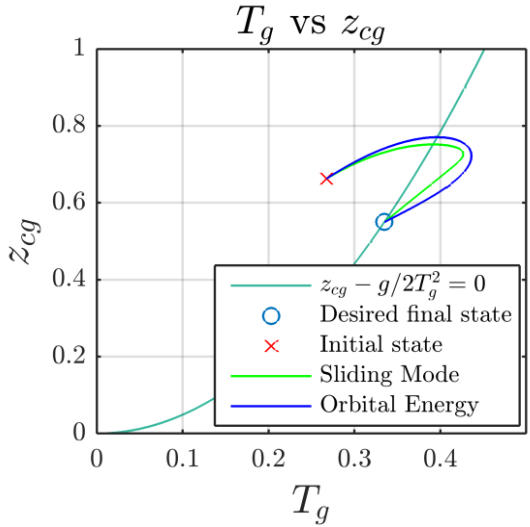

Figure 12: Control of $T_g$ and $z_{cg}$ in the case of a disturbance rejection.

### H. Flat foot orientation transformation.

In this subsection we will show how to transform and apply analysis and control showed in subsections *E*, *F* and *G*. Recalling (6):

$$\ddot{r} = u(r - r_P) + g, u \geq 0, Ar_P \leq b$$

All feasible points $r_P$ belongs to a plane containing the flat foot. If $\hat{n}$ is the normal vector of the foot, the following equality is hold for some constant $c_P$:

$$\hat{n}^T r_P = c_P$$

Instead of using the $z$-component of $r$, we will use:

$$z_r = \frac{\hat{n}^T r - c_P}{\hat{n}^T \hat{e}_z} \quad (92)$$

This transforms the dynamics of (6) to:

$$\begin{bmatrix} \ddot{r}_{xy} = u(r_{xy} - r_{Pxy}) \\ \ddot{z}_r = u(z_r - 0) - g \end{bmatrix}$$

Which can be expressed in compact form as:

$$\ddot{r}_r = u(r_r - r_{rP}) + g, u \geq 0, A_r r_{rP} \leq b_r \quad (93)$$

With $\hat{e}_z^T r_{rP} = z_P = 0$. So transformation (92) rotates any orientation of the foot to a horizontal one. Matrix $A$ and vector $b$ changed to $A_r$ and $b_r$ because of the use of $z_r$ instead of $z$. Note that this rotation must be performed every time the Contact Surface is changed, i.e., every time the robot steps. Simulations were done using feedback over this "normalized" orientation of the foot (93). Fig. 7-12 are plotted in the original frame of coordinates, so analysis and control showed in subsections *E*, *F* and *G* are still valid in a non-horizontal Contact Surface after the linear change of variables (92).

## VI. DISCUSSION

We have given some conditions and requirements for the 3D VHIP with 0-step capturability, but more work should be done. We are not taking into account the kinematic constraints: When the foot is close to the ballistic trajectory itself (low values of $z_{cg}$) the robot tends to let the CoM fall and CoM doesn't stop falling until the very last moment. In a similar way, when the foot is close to the vertical projection of the CoM (low values of $T_g$), the robot tends to apply an extremely high height variation. Both are kinematic problems: in the first case the robot will crash into the ground, and in the second case the robot will jump. Actuation limits are also a big issue, although we have used the feedback controller from [15] considering $u \leq u_{max}$, it is not limiting directly the force. Friction limits are also not taken into account. Although friction apparently is not so difficult to handle in the single contact case (the ballistic trajectory eventually enters to the friction cone of the foot, except in extreme cases), it is a hard case to solve when it is combined with limited actuation. Also, the multi-contact friction case is much more complicated.

An interesting topic of research is the extension of these analytical methods and feedback based controls to the walking problem. One approach to solve this is to study the N-step capturability, and perform continuously a 1-step or 2-step capturability. Another approach is to predefine the DCM trajectory and find feedback control laws for tracking this reference. Although a TV-LQR control from [21] can be used straightforward in a linearization around the trajectory, it is interesting to find control laws that are able to track the DCM trajectories in the whole possible region of stability.

A last open problem is the inclusion of a variable angular momentum. This then requires the control of the Full Centroidal Dynamics, with this work being an inner approximation where the equation $\dot{L} = 0$ is held.

## VII. CONCLUSION

In conclusion, with either a fixed or variable CoP, the robot always must step in the ballistic line to be 0-step capturable. Otherwise, the robot will fall or it will need to use n-step recovery.

In any case, the robot can just Fix the CoP in a place below the ballistic trajectory and it can perform a 2D VHIP strategy for recovery in the ballistic plane.

If the robot wants to make use of the Variable CoP, the motion can be decomposed into two different analyses: one more intuitive using the gDCM, or another splitting that divides the analysis into height strategies and CoP strategies. We also provide a way to reinitialize the Augmented 3D VHIP in the event that a disturbance produces the robot to fall outside the region of stability. Finally, we provided a small generalization of the bounded case $u \leq u_{max}$ and we show that the present analysis works in any orientation of the foot using a linear transformation to become it horizontal.